\documentclass{article} % For LaTeX2e
\usepackage{nips13submit_e,times}

\usepackage{makeidx}  % allows for indexgeneration
\usepackage{booktabs}
\usepackage[font=footnotesize]{caption}
\usepackage{subfigure}
\usepackage{comment}
\usepackage{hyperref}
\usepackage{url}
\usepackage{sidecap}
\sidecaptionvpos{figure}{c}
\usepackage{multicol}
\usepackage{enumitem}

\usepackage{amsmath}
\usepackage{amssymb}
\usepackage{mathtools}

\usepackage{algorithm}
\usepackage{algorithmic}

\usepackage{multirow}
\usepackage{pbox}
\usepackage{array}
\newcolumntype{L}[1]{>{\raggedright\let\newline\\\arraybackslash\hspace{0pt}}m{#1}}
\newcolumntype{C}[1]{>{\centering\let\newline\\\arraybackslash\hspace{0pt}}m{#1}}
\newcolumntype{R}[1]{>{\raggedleft\let\newline\\\arraybackslash\hspace{0pt}}m{#1}}

\usepackage[T1]{fontenc}

\title{A Bayesian Tensor Factorization Model via Variational Inference for Link Prediction}

\author{
Beyza Ermi\c{s}  \\
Department of Computer Engineering \\
Bo\u{g}azi\c ci University, Turkey \\
\texttt{beyza.ermis@boun.edu.tr} 
\And
A. Taylan Cemgil \\
Department of Computer Engineering \\
Bo\u{g}azi\c ci University, Turkey \\
\texttt{taylan.cemgil@boun.edu.tr}
}

\nipsfinalcopy

\begin{document}

\maketitle

\begin{abstract}
Probabilistic approaches for tensor factorization aim to extract meaningful structure from incomplete data by postulating low rank constraints. Recently, variational Bayesian (VB) inference techniques have successfully been applied to large scale models. 
This paper presents full Bayesian inference via VB on both single and coupled tensor factorization models. Our method can be run even for very large models and is easily implemented. It exhibits better prediction performance than existing approaches based on maximum likelihood on several real-world datasets for missing link prediction problem.
\end{abstract}

%% ----------------------------------------------------------------------
%% M A I N   B O D Y
%% ----------------------------------------------------------------------

\section{Introduction}
\label{sec:intro}

Factorization based data modelling has become popular together with the advances in the computational power. Non-negative Matrix Factorization (NMF) model, proposed by Lee and Seung \cite{Lee_Seung_1999} (and also earlier by Paatero and Tapper \cite{PaTa94}), is one of the most popular factorization models. Tensors are defined as a natural generalization of matrix factorization, when observed data have several semantically meaningful dimensions. These modelling paradigms have found place in many fields including recommender systems \cite{ZhCaZhXiYa10}, image processing \cite{cemgil09-nmf} and bioinformatics \cite{cichocki09}. Typically, the objective function in matrix and tensor factorization problems could be minimized by a suitable optimization algorithm. Here, we solve this problem by using a probabilistic approach based on variational Bayes (VB) that provides better prediction performance and scales to very large datasets.

The Probabilistic Latent Tensor Factorization framework (PLTF) \cite{Yilmaz:2010} is appeared as an extension of the matrix factorization model~\cite{Lee_Seung_1999} and enables one to incorporate domain specific information to any arbitrary factorization model and provides the update rules for multiplicative expectation-maximization (EM) algorithms. In this framework, the goal is to compute an approximate factorization of a given higher-order tensor, i.e., a multiway array, $X$ in terms of a product of individual factors $Z_\alpha$ as:
\vspace*{-2mm}
\begin{align}
	X(v_0)\approx\hat{X}(v_0)=\sum_{\bar{v}_0}\prod_\alpha Z_\alpha(v_\alpha), \label{eqn:tensorFact}
\end{align}
where some of the factors are possibly fixed. Here, we define $v$ as the set of all indices in a model, $v_0$ as the set of visible indices, $v_\alpha$ as the set of indices in $Z_\alpha$, and $\bar{v}_\alpha = v- v_\alpha$ as the set of all indices not in $Z_\alpha$ and $\alpha = 1,... K$ as the factor index. Since the product $\prod_\alpha Z_\alpha(v_\alpha)$ is collapsed over a set of indices, the factorization is latent.
  
The Generalized Coupled Tensor Factorization (GCTF)~\cite{yilmazGTF} model takes the PLTF model one step further where, in this case, we have multiple observed tensors $X_\nu$ that are supposed to be factorized simultaneously:

\noindent\begin{minipage}[b]{0.50\textwidth}
\vspace*{-3mm}
\begin{align}
	X_\nu(v_{0,\nu}) \approx \hat{X}_\nu(v_{0,\nu}) = \sum_{\bar{v}_{0,\nu}}  \prod_\alpha Z_\alpha(v_\alpha)^{R^{\nu,\alpha}} \label{eqn:gctf}
\end{align}
\end{minipage}
\noindent\begin{minipage}[b]{0.50\textwidth}
\vspace*{-3mm}
\begin{align}
  R^{\nu,\alpha} &= \left\{
   \begin{array}{l l}
      1 &  \quad \text{if $X_\nu$ and $Z_\alpha$ connected}\\
      0 &  \quad \text{otherwise}  \\
   \end{array} \right. .
\label{eqn:Rmatrix}
\end{align}
\end{minipage}

where $\nu = 1,... |\nu|$ and $R$ is a {\it coupling matrix} that is defined as in (\ref{eqn:Rmatrix}).
Note that, distinct from PLTF model, there are multiple visible index sets ($v_{0,\nu}$) in the GCTF model, each specifying the attributes of the observed tensor $X_\nu$. 
 
In this study, to model a multiway data, we use non-negative variants of the two most widely-used low-rank tensor factorization models; the Tucker model \cite{Tu66} and the more restricted CANDECOMP/PARAFAC (CP) model \cite{Ha70,CaCh70,Hi27a}. In order to illustrate the approach, we can define these models in the PLTF notation. Given a three-way tensor $X$ the CP model is defined as follows:
\vspace*{-2mm}
\begin{align}
    X(i,j,k) &\approx\hat{X}(i,j,k)=\sum_r Z_1(i,r)Z_2(j,r)Z_3(k,r) \label{eq:CP}
\end{align}
where the index sets $v=\{i,j,k,r\}$, $v_0=\{i,j,k\}$, $v_1=\{i,r\}$, $v_2=\{j,r\}$ and $v_3=\{k,r\}$. In addition, for coupled analysis of relational datasets represented as heterogeneous data, we model the data by using GCTF notation and simultaneously fit a large class of tensor models to higher-order tensors/matrices with common latent factors .
%An alternative Tucker model of $X$ is defined in the PLTF notation as follows:
%\begin{align}
%    \hat{X}(i,j,k) &=\sum_{p,q,r} Z_1(i,p)Z_2(j,q)Z_3(k,r)Z_4(p,q,r) \label{eq:Tucker2}
%\end{align}

%Relational learning can be used to augment one data source with other correlated sources of information, to improve predictive accuracy. We frame relational learning problems as tensor factorization problems, and propose a variational Bayesian model.
Our main contribution in this paper is a novel variational Bayesian procedure for making inference on the PLTF and GCTF frameworks. In this method, the exact characterization of the approximating distribution and full conditionals are observed as a product of multinomial distributions, leading to a richer approximation distribution than a naive mean field. 
Our method can be formulated entirely in terms of sparse and low rank tensors, and it is easily scaled up to very large problem.
We illustrate the proposed approach on large-scale link prediction problem: the problem of predicting the existence of connections between entities of interest. 

\subsection{Preliminaries}
\paragraph{\textbf{Probability Model:}} The usual approach to estimate the factors $Z_\alpha$ is trying to find the optimal 
$Z^{*}_{1:K} = {\operatorname{argmin}}_{Z_{1:K}} \hspace{1mm} d(X || \hat{X})$, where $d(.)$ is a divergence typically taken as Euclidean, Kullback-Leibler or Itakura-Saito divergences. Since the analytical solution for this problem is intractable, one should refer to iterative or approximate inference methods. 

\noindent\begin{minipage}[b]{0.40\textwidth}
\centering
    \includegraphics[scale=0.9]{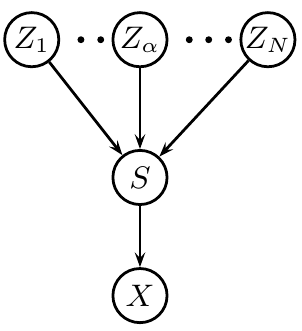}
    \captionof{figure}{The generative model of the PLTF framework as a Bayesian network. The directed acyclic graph describes the dependency structure of the variables: the full joint distribution can be written as $p(X,S,Z_{1:K}) = p(X|S)p(S|Z_{1:K})\prod_\alpha p(Z_\alpha)$.}
    \label{fig:bayesianNetwork}
\end{minipage}
\noindent\begin{minipage}[b]{0.60\textwidth}
\small
\begin{flalign}
  \tag{factor priors}
  &Z_\alpha(v_\alpha) \sim \mathcal{G}(Z_\alpha(v_\alpha); A_\alpha(v_\alpha), B_\alpha(v_\alpha))   \label{eqn:factorPriors} \\
  \tag{intensity}  
  &\Lambda(v) = \prod_\alpha Z_\alpha(v_\alpha) \label{eqn:intensity} \\
  \tag{components}
  &S(v) \sim \mathcal{PO}(S(v); \Lambda(v)) \label{eqn:components}  \\
  \tag{observation}
  &X(v_0) = \sum_{\bar{v}_0} S(v) \label{eqn:observation} \\
  \tag{parameter}
  &\hat{X}(v_0) = \sum_{\bar{v}_0} \Lambda(v) \label{eqn:parameter}  \\
  \tag{poisson}
  &\mathcal{PO}(s;\lambda) = e^{-\lambda} \frac{\lambda^{s}}{s!}	\label{eqn:poisson_dist} \\
  \tag{gamma}
  &\mathcal{G}(z;a,b) = e^{-bz}	\frac{z^{a-1}b^{a}}{\Gamma(a)}. \label{eqn:gamma_dist}
\end{flalign}
\end{minipage}
\normalsize

In this study, we use the Kullback-Leibler (KL) divergence as the cost function which is equivalent to selecting the Poisson observation model \cite{cemgil09-nmf,Yilmaz:2010}, while our approach can be extended to other costs where a composite structure is present. The overall probabilistic model where the symbols refer to Poisson and Gamma distributions respectively is defined as above.
The Gamma prior on the factors are chosen in order to preserve conjugacy. The graphical model for the PLTF framework is depicted in 
Figure~\ref{fig:bayesianNetwork}. 
Note that $p(X|S)$ is a degenerate distribution that is defined as: $ p(X|S) =  \prod_{v_0} \delta \big(X(v_0) - \sum_{\bar{v}_0} S(v) \big)$.
Here, $\delta(.)$ is the Kronecker delta function where $\delta(x) = 1$ when $x = 0$ and $\delta(x) = 0$ otherwise.

\paragraph{\textbf{Missing data:}} To model missing data, we define a 0-1 mask array $M$, the same size as $X$ where $M(v_{0})=1$ ($M(v_{0})=0$) if $X(v_{0})$ is observed (missing). Using the mask variables, the missing data is handled smoothly by the following observation model in PLTF:
\begin{align}
p(X|S)p(S|Z_{1:N}) = \prod_{v_0}\prod_{\bar{v}_0}\lbrace p(X(v_0)|S(v))p(S(v)|Z_{1:N})\rbrace^{M(v_0)}
\label{eqn:handle_missing}
\end{align} 

\paragraph{\textbf{Tensor forms via $\Delta$ function:}} We make use of $\Delta$ function to make the notation shorter and implementation friendly. A tensor valued $\Delta_{\alpha}(Q)$ function associated with component $Z_\alpha$ is defined as follows:
\small
\begin{align}
  \Delta_{\alpha}(Q) = \big[ \sum_{\bar{v}_\alpha} \big( Q(v_0) \prod_{\alpha\prime\neq\alpha} Z_\alpha\prime(v_\alpha\prime)\big) \big]    \label{eqn:delta_func} 
\end{align}
\normalsize
Recall that $\Delta_{\alpha}(Q)$ is an object the same size of $Z_\alpha$ while $\Delta_{\alpha}(Q)(v_\alpha)$ refers to a particular element of $\Delta_{\alpha}(Q)$.

\paragraph{\textbf{Fixed Point Update Equation for PLTF and GCTF:}} Here, we recall the generative PLTF model with KL loss defined in (\ref{eqn:factorPriors}) with the following fixed point iterative update equation for the component $Z_{\alpha}$ obtained via EM as:
\small
\begin{align}
Z_{\alpha}(v_{\alpha}) \leftarrow \frac{(A_{\alpha}(v_{\alpha})-1)+Z_{\alpha}(v_{\alpha})\sum_{\bar{v}_{\alpha}}M(v_0)\frac{X(v_0)}{\hat{X}(v_0)}\prod_{\alpha\prime\neq\alpha}Z_{\alpha\prime}(v_{\alpha\prime})}{\frac{A_{\alpha}(v_{\alpha})}{B_{\alpha}(v_{\alpha})}+\sum_{\bar{v}_{\alpha}}M(v_0)\prod_{\alpha\prime\neq\alpha}Z_{\alpha\prime}(v_{\alpha\prime})}
\label{eqn:updateEq_EM}
\end{align}
\normalsize 
where $\hat{X}(v_0)$ is the model estimate defined as earlier $\hat{X}(v_0)=\sum_{\bar{v}_{0}}\prod_{\alpha}Z_{\alpha}(v_{\alpha})$. We note that the gamma hyperparameters $A_\alpha(v_\alpha)$ and $B_\alpha(v_\alpha)/A_\alpha(v_\alpha)$ are chosen for computational convenience for sparseness representation such that the distribution has a mean $B_{\alpha}(v_{\alpha})$ and standard deviation $B_{\alpha}(v_{\alpha})/\sqrt{A_{\alpha}(v_{\alpha})}$ and for small $A_{\alpha}(v_{\alpha})$ most of the parameters are forced to be around 0 favoring for a sparse representation \cite{cemgil09-nmf}. So, Equation(\ref{eqn:updateEq_EM}) can be approximated and written into a form that by use of $\Delta_{\alpha}^{Z}(.)$ as: 
\vspace*{-1mm}
\begin{align}
	Z_\alpha \leftarrow Z_\alpha \circ \Delta_{\alpha} (M \circ X/\hat{X})/\Delta_{\alpha}(M)  \label{eqn:updateEq_EM2}
\end{align}
where as usual $\circ$ and $/$ stand for element wise multiplication(Hadamard product) and division respectively. In addition, one can obtain the following compact fixed point equation for the update of $Z_\alpha$ in GCTF model with KL loss:
\begin{align}
	Z_\alpha \leftarrow Z_\alpha \circ \sum_\nu R^{\nu,\alpha} \Delta_{\alpha} (M \circ X/\hat{X})/ \sum_\nu R^{\nu,\alpha} \Delta_{\alpha}(M)  \label{eqn:updateEq_EM3}
\end{align}
We use update equation (\ref{eqn:updateEq_EM2}) for PLTF-EM method to compare with the PLTF-VB method and (\ref{eqn:updateEq_EM3}) for GCTF-EM method to compare with the GCTF-VB method in the following chapters.

\section{Variational Bayes}
\label{sec:method}

For a Bayesian point of view, a model is associated with a random variable $\Theta$ and interacts with the observed data X simply as $p(\Theta|X) \propto p(X|\Theta)p(\Theta)$. The quantity $p(X|\Theta)$ is called \textit{marginal likelihood} \cite{bishop06} and it is average over the space of the parameters, in our case, $S$ and $Z$ as \cite{cemgil09-nmf}.
\begin{align}
	p(X|\Theta) = \int_Z dZ \sum_S p(X|S,Z,\Theta) p(S,Z|\Theta) \label{eqn:marginal_lh}
\end{align}

On the other hand, computation of this integral is itself a difficult task that requires averaging on several models and parameters. There are several approximation methods such as sampling or deterministic approximations such as Gaussian approximation. One other approximation method is to bound the log marginal likelihood by using \textit{variational inference} \cite{cemgil09-nmf,bishop06,GhahramaniB00} where an approximating distribution $q$ is introduced into the log marginal likelihood equation:
\begin{align}
	\log p(X|\Theta) \geq \int_Z dZ \sum_S q(S|Z) \log \frac{p(X,S,Z|\Theta)}{q(S,Z)}   \label{eqn:app_dist}
\end{align}
where the bound attains its maximum and becomes equal to the log marginal likelihood whenever $q(S,Z)$ is set as $p(S,Z|X,\Theta)$, that is the exact posterior distribution. However, the posterior is usually intractable, and rather, inducing the approximating distribution becomes easier. Here, the approximating distribution $q$ is chosen such that it assumes no coupling between the hidden variables such that it factorizes into independent distributions as $q(S,Z) = q(S) q(Z)$. As exact computation is intractable, we will resort to standard variational Bayes approximations \cite{bishop06,GhahramaniB00}. The interesting result is that we get a belief propagation algorithm for marginal intensity fields rather than marginal probabilities.

\subsection{Variational Update Equation for Probabilistic Tensor Factorization (PLTF-VB)}
\label{sec:var_update_eq}
Here, we formulate the fixed point update equation for the update of the factor $Z_{\alpha}$ as an expectation of the approximated posterior distribution \cite{YKYthesis}. Approximation for posterior distribution $q(Z)$ is identified as the gamma distribution with the following parameters:

\small
\hspace*{-2mm}
\noindent\begin{minipage}[b]{0.50\textwidth}
\begin{align}
	C_\alpha(v_\alpha) &= A_\alpha(v_\alpha) + \sum_{\bar{v}_{\alpha}} M(v_0) \frac{X(v_0)}{\hat{X}_L(v_0)} \prod_{\alpha} L_{\alpha}(v_{\alpha}) \label{eqn:shapePar} 
\end{align}
\end{minipage}	
\hspace*{3mm}
\noindent\begin{minipage}[b]{0.50\textwidth}
\begin{align}		
	D_\alpha(v_\alpha) &= \big( \frac{A_{\alpha}(v_\alpha)}{B_{\alpha}(v_\alpha)} + \sum_{\bar{v}_{\alpha}} M(v_0) \prod_{\alpha\prime\neq\alpha} \langle Z_{\alpha\prime}(v_{\alpha\prime}) \rangle \big)^{-1}  \label{eqn:scalePar}
\end{align} 
\end{minipage}
\normalsize
Hence the expectation of the factor $Z_{\alpha}$ is identified as the mean of the gamma distribution and given in the iterative fixed point update equation obtained via variational Bayes:
\small
\begin{align}
\langle Z_{\alpha}(v_{\alpha}) \rangle &= C_\alpha(v_\alpha) D_\alpha(v_\alpha) \label{eqn:IFPUE}  \\   
&= \frac{A_{\alpha}(v_{\alpha})+L_{\alpha}(v_{\alpha})\sum_{\bar{v}_{\alpha}} M(v_0) \frac{X(v_0)}{\hat{X}_L(v_0)} \prod_{\alpha\prime\neq\alpha} L_{\alpha\prime}(v_{\alpha\prime})}{\frac{A_{\alpha}(v_{\alpha})}{B_{\alpha}(v_{\alpha})} + \sum_{\bar{v}_{\alpha}} M(v_0) \prod_{\alpha\prime\neq\alpha} E_{\alpha\prime}(v_{\alpha\prime})} \label{eqn:updateEq_VB}
\end{align}
\normalsize 
$E_{\alpha}(v_{\alpha})$ and $L_{\alpha}(v_{\alpha})$ ($L$ due to `Log') are two forms of expectations of $Z_{\alpha}(v_{\alpha})$ while $\hat{X}_E(v_0)$ and $\hat{X}_L(v_0)$ are model outputs generated by the components $E_{\alpha}(v_{\alpha})$ and $L_{\alpha}(v_{\alpha})$. While $\hat{X}_E$ is not being used in Equation(\ref{eqn:updateEq_VB}) we define it here, in addition to $\hat{X}_L$, (and use it later on) since $\hat{X}_E$ has the same shape as $\hat{X}_L$. Indeed $\hat{X}_E$ and $\hat{X}_L$ can be regarded as different `views' of $\hat{X}$ since they have the same shape (dimensions) as $\hat{X}$ and their computations are done via the same matrix primitives as $\hat{X}$. Here:
\vspace*{-2mm}
\hspace*{-10mm}
\noindent\begin{minipage}[b]{0.60\textwidth}
\begin{align}
	E_{\alpha}(v_{\alpha}) &= \langle Z_{\alpha}(v_{\alpha}) \rangle = C_{\alpha}(v_{\alpha}) D_{\alpha}(v_{\alpha})   \label{eqn:expE}  \\
	\vspace*{2mm}
	L_{\alpha}(v_{\alpha}) &= \exp\big(\langle \log Z_{\alpha}(v_{\alpha}) \rangle \big) \notag \\
	                                 &= \exp \big( \psi \big(C_{\alpha}(v_{\alpha}) \big) \big) D_{\alpha}(v_{\alpha})   \label{eqn:expL}  
\end{align} 
\end{minipage}
\vspace*{2mm}
\noindent\begin{minipage}[b]{0.40\textwidth}
\begin{align}
	\hat{X}_E(v_0) &= \sum_{\bar{v}_0} \prod_{\alpha} E_{\alpha}(v_{\alpha})   \label{eqn:XexpE}  \\
	\hat{X}_L(v_0) &= \sum_{\bar{v}_0} \prod_{\alpha} L_{\alpha}(v_{\alpha})   \label{eqn:XexpL}  
\end{align} 
\end{minipage}

Note that the VB version of the update equation (\ref{eqn:updateEq_VB}) closely resembles the EM version (PLTF-EM) given in Equation~\ref{eqn:updateEq_EM}.
Indeed when the observed values are large, digamma function becomes $\lim_{x\rightarrow\infty}\psi(x)/\log(x)=1$, and this, in turn, gives $L_\alpha(v_\alpha) \simeq E_\alpha(v_\alpha)$ and $\hat{X}_L(v_0) \simeq \hat{X}_E(v_0)$.

\subsection{Variational Update Equation for Coupled Tensor Factorization (CTF-VB)}
\label{sec:VB_CTF}

Here, we present a variational Bayesian method to make inference on the coupled tensor factorization models and to derive update equations for these models that handles the simultaneous tensor factorizations where multiple observations tensors are available.
We present variational Bayesian coupled tensor factorization as an approach to exploiting side information, i.e., each decomposition is coupled by sharing some factor matrices. We use this method to improve the performance of PLTF-VB algorithm defined in Section~\ref{sec:var_update_eq} by incorporating knowledge in the additional matrices. 

In this case, we address the problem when multiple observed tensors $X_\nu$ for $\nu = 1 . . . |\nu|$ are factorised simultaneously. Each observed tensor $X_\nu$ now has a corresponding index set $v_{0,\nu}$ and a particular configuration will be denoted by $v_{0,\nu} \equiv u_\nu$. And, we also define the $|\nu| \times |\alpha|$ coupling matrix $R$. Finally, we define another particular configuration for index set of $S_\nu$ will be denoted by $\bigcup_{R_{\nu,\alpha}=1} v_\alpha \equiv r_\nu$.

For PLTF-VB method, we obtain the approximating distributions as:
\begin{align}
	q_{S(v_0,:)} \sim \mathcal{M}(S(v_0,:), X(v_0), P(v_0,:))  \notag 
\end{align}
where the cell probabilities and sufficient statistics for $q_{S(v_0,:)}$ are: 
\begin{align}
	P(v) = \frac{\prod_\alpha L_\alpha(v_\alpha)}{\sum_{\bar{v}_0} \prod_\alpha L_\alpha(v_\alpha)}  \notag  \hspace*{20mm}
	\langle S(v)\rangle = X(v_0) P(v)    \notag
\end{align}

For the coupled factorization (CTF-VB), we get the following expression of the cell probabilities $P_\nu(r_\nu)$ here as:
\begin{align}
	P_\nu(r_\nu) = \frac{\exp(\sum_\alpha \langle \log Z_\alpha(v_\alpha)\rangle)}{\sum_{\bar{u}_\nu} \exp(\sum_\alpha\langle \log Z_\alpha(v_\alpha)\rangle)}   
= \frac{\prod_\alpha \exp(\langle\log Z_\alpha(v_\alpha) \rangle)}{\sum_{\bar{u}_\nu} \prod_\alpha \exp(\langle\log Z_\alpha(v_\alpha) \rangle)}   
= \frac{\prod_\alpha L_\alpha(v_\alpha)}{(\hat{X}_L)_{v}(u_\nu)}  \label{eqn:cell_prob_compact_coupled3}
\end{align}
Then the sufficient statistics $\langle S_\nu(r_\nu) \rangle$ turns to 
\begin{align}
    \langle S_v(r_\nu) \rangle = X_\nu(u_\nu)P_v(r_\nu) = \frac{X_v(u_\nu)}{(\hat{X}_L)_{v}(u_\nu)}\prod_\alpha L_\alpha(v_\alpha)   \label{eqn:suff_stat_coupled}
\end{align}

Now we turn to formulating $q(Z)$. The distribution $q_{Z_\alpha(v_\alpha)}$ is obtained similarly as after we expand the log and drop irrelevant terms it becomes proportional to 
\begin{align}
q_{Z_{\alpha}(v_\alpha)} &\propto \exp \big( \langle \log p(S|Z) + \log p(Z|\Theta)\rangle_{q/q_{Z_{\alpha}(v_\alpha)}} \big) 
\propto \log Z_\alpha(v_\alpha) \big( A_\alpha(v_\alpha) -1  \label{eqn:qz3} \\  
& + \sum_\nu R^{\nu,\alpha} \sum_{\bar{v}_\alpha}\langle S_\nu(r_\nu)\rangle^{R^{\nu,\alpha}} \big) - Z_\alpha(v_\alpha) \big( \frac{A_\alpha(v_\alpha)}{B_\alpha(v_\alpha)} + \sum_\nu R^{\nu,\alpha} \sum_{\bar{v}_\alpha} \prod_{\alpha\prime\neq\alpha} \langle Z_{\alpha\prime}(v_{\alpha\prime})\rangle^{R^{\nu,\alpha}} \big) \notag
\end{align}
which is the distribution $q_{Z_{\alpha}(v_\alpha)}  \sim \mathcal{G}(C_\alpha(v_\alpha), D_\alpha(v_\alpha))$
where the shape and scale parameters for $q_{{Z_\alpha(v_\alpha)}}$ are
\small
\begin{align}
	C_\alpha &= A_\alpha + \sum_\nu R^{\nu,\alpha} L_\alpha \circ \Delta_{\alpha}(M_\nu \circ X_\nu /(\hat{X}_L)_{\nu})    \label{eqn:hyperC_final_coupled}  \\
	D_\alpha &= \big(A_\alpha / B_\alpha + \sum_\nu R^{\nu,\alpha} \Delta_{\alpha}(M_\nu) \big)^{-1}  \label{eqn:hyperD_final_coupled} 
\end{align}
\normalsize
that, in turn, since $\langle Z_\alpha\rangle$ is $C_\alpha \circ D_\alpha$, the sufficient statistics for $q(Z_\alpha)$ become:
\begin{align}
\langle Z_\alpha\rangle &= E_\alpha \leftarrow \frac{A_\alpha + \sum_\nu R^{\nu,\alpha} L_\alpha \circ \Delta_{\alpha}(M_\nu \circ X_\nu /(\hat{X}_L)_{\nu})}{\frac{A_\alpha}{B_\alpha} + \sum_\nu R^{\nu,\alpha} \Delta_{\alpha}(M_\nu)}    
\label{eqn:Z_final_coupled}  
\end{align}

\section{Experiments and Results}
\label{sec:experiments}

%In this section, we demonstrate the use of the proposed variational Bayesian coupled tensor factorization method (CTF-VB) for model selection and missing link prediction.
%First, we study model selection and show that the proposed approach can accurately determine the number of components in a CP model. We also study the missing link prediction problem and show the performance of CTF-VB for model selection on two real data sets., i.e., the UCLAF~\cite{ZhCaZhXiYa10} and Digg~\cite{LiSuCaKo09}. We first demonstrate that coupled tensor factorizations (CTF-VB) outperform low-rank approximations of a single tensor (PLTF-VB), then we compare the performance of the proposed variational Bayesian approach (CTF-VB) with the standard approach (CTF-EM) in terms of missing link prediction recovery.
%For the experiments we use the algorithm that implements variational fixed point update equation given in panel Algorithm~\ref{alg:pltf_vb} and we use the equation given in (\ref{eqn:bound2}) for variational bound computation.

In this section, we demonstrate the use of the proposed variational Bayesian coupled tensor factorization method (GCTF-VB) for missing link prediction problem in order to show that joint analysis of data from multiple sources via coupled factorization significantly improves the link prediction performance. We evaluate the performance of GCTF-VB on two real data sets., i.e., the UCLAF~\cite{ZhCaZhXiYa10} and Digg~\cite{LiSuCaKo09}. First, we demonstrate that coupled tensor factorizations (GCTF-VB) outperform low-rank approximations of a single tensor (PLTF-VB), then we compare the performance of the proposed variational Bayesian approach (GCTF-VB) with the standard approach (GCTF-EM) in terms of missing link prediction recovery. For the experiments, we use the algorithm that implements variational fixed point update equations given in Equation~\ref{eqn:updateEq_VB} (for PLTF-VB) and Equation~\ref{eqn:Z_final_coupled} (for GCTF-VB). 

% UCLAF
\paragraph*{\textbf{Data:}} As the small scale real data, we use the UCLAF dataset \footnote{\url{http://www.cse.ust.hk/~vincentz/aaai10.uclaf.data.mat}} extracted from the GPS data that include information of three types of entities: user, location and activity. The relations between the user-location-activity triplets are used to construct a three-way tensor $X_1$. 
In tensor $X_1$, an entry $X(i,j,k)$ equals 1 if user $i$ visits location $j$ and performs activity $k$ there, otherwise it equals 0. 
\begin{comment}
In tensor $X_1$, an entry $X(i,j,k)$ indicates the frequency of a user $i$ visiting location $j$ and doing activity $k$ there; otherwise, it is $0$. Since we address the link prediction problem in this study, we define the user-location-activity tensor $X_1$ as:
 \begin{align}
  X(i,j,k) = 
   \begin{cases}
      1 & \text{if user $i$ visits location $j$ and performs activity $k$ there,} \\
      0 & \text{otherwise.}
   \end{cases} 
\end{align}
\end{comment}
The collected data also includes additional side information: the user-location preferences from the GPS trajectory data and the location features from the POI (points of interest) database, represented as the matrix $X_{2}$ and $X_{3}$ respectively. Our aim is to restore the missing links in ${X}_1$, by using the three-way observation tensor ${X}_1$ and two auxiliary matrices ${X}_2$ and ${X}_3$ that provide side information. This is a difficult link prediction problem since ${X}_1$ contains less than 1\% of all possible links or an entire slice of ${X}_1$ may be missing. Note that, the number of users is $146$, the number of locations $168$, the number of activities $5$ and number of location features $14$ in our experiments.

% Digg
Furthermore, we address link prediction problem on a large-scale dataset\footnote{\url{http://www.public.esu.edu/~ylin56/kdd09sup.html}} collected from Digg in order to show the scalability of the proposed approach. Digg is a social news resource that allows users to submit, digg and comment on news stories. Lin \emph{et al.}~\cite{LiSuCaKo09} have collected data from a large set of user actions from Digg. It includes stories, users and their actions (submit, digg, comment and reply) with respect to the stories, as well as the explicit friendship (contact) relation among these users. It also includes the topics of the stories and keywords extracted from the titles of stories. There are five types of entities: user, story, comment, keyword and topic and six relationships among them (see ~\cite{LiSuCaKo09} for a comprehensive illustration of relations).

We will use three relationships in this study: user-story-comment (R1), story-keyword-topic (R2) and user-story (R3). We represent each relation with a tensor with sizes $9583 \times 44005 \times 241800$, $44005 \times 13714 \times 51$ and $9583 \times 44005$ and the total number of tuples in each integrated data tensor per relation is $151.779$, $1.157.529$ and $94.551$ respectively. The prediction results are compared with the actual diggs and comments as ground truth.
Based on the Digg scenario, we design two prediction tasks on Digg dataset: (i) comment prediction - what stories a user will comment on, (ii) digg prediction - what stories a user will digg.

For each prediction task, we form different coupled models and solve these models with proposed approach. Table~\ref{table:CoupledModels} includes some of these models for both datasets.
\begin{table}[h!]
\begin{center}
\scalebox{0.80}{
\renewcommand{\arraystretch}{1.3}
\begin{tabular}{| l | l |}
%\hline   \multicolumn{2}{| c |}{Coupled CP Model}    \\ \cline{1-2}
\hline UCLAF Dataset  &   Digg Dataset (Comment Prediction)  \\ \hline \\ [-2.5ex]
$\begin{aligned} \hat{X}_1({i,j,k}) &= \sum_r A({i,r}) B({j,r}) C({k,r})  \\ 
	                      \hat{X}_2({i,m}) &= \sum_r  A({i,r}) D({m,r}) \\   
	                      \hat{X}_3({j,n}) &= \sum_r   B({j,r}) E({n,r})   \end{aligned}$ &                               
$\begin{aligned} \hat{X}_1({i,j,k}) &= \sum_r A({i,r}) B({j,r}) C({k,r}) \\
                          \hat{X}_2({j,m,n}) &= \sum_r  B({j,r}) D({m,r}) E({n,r})  \end{aligned}$ \\ \hline                                     
\end{tabular}}
\caption{Example CP-coupled tensor factorization models for link prediction on  UCLAF and Digg datasets.}
\label{table:CoupledModels}
\end{center}
\vspace*{-5mm}
\end{table}
%\paragraph*{\textbf{Computational Environment:}} All experiments were performed using MATLAB 2010b on 2.4GHz Core i5 520M processor and 4GB RAM. Timings were performed using MATLAB's tic and toc functions.
\paragraph{\textbf{Computational Complexity:}} Assuming that all datasets have equal number of dimensions, i..e, a tensor is an $N \times N \times N$ array while the coupled matrix is of size $N \times N$, then the leading term in the computational complexity of the coupled model will be due to the updates for the tensor model. For an $R$-component CP model, for instance, that would be $O(N^3R)$.

If a large number of entries is missing, then mask tensor $M$ is sparse. In this case, there is no need to allocate storage for every entry of the tensor $X$. Instead, we can store and work with just the known values, making the method efficient in both storage and time.
Our approach also has ability to perform sparse computations, enabling it to scale to very large real datasets using specialized sparse data structures, significantly reducing the storage and computation costs. When we take into account the sparsity pattern of the data, the time complexity of each iteration is roughly $O(N)$, which is linear in terms of the total number of non-missing entries $N$.

\begin{figure}[h!]
\hspace*{-4mm}
\begin{minipage}[b]{0.55\textwidth}
\centering
\subfigure[$500 \times 500 \times 500$]{\label{subfig:scalable500}\includegraphics[scale=0.40]{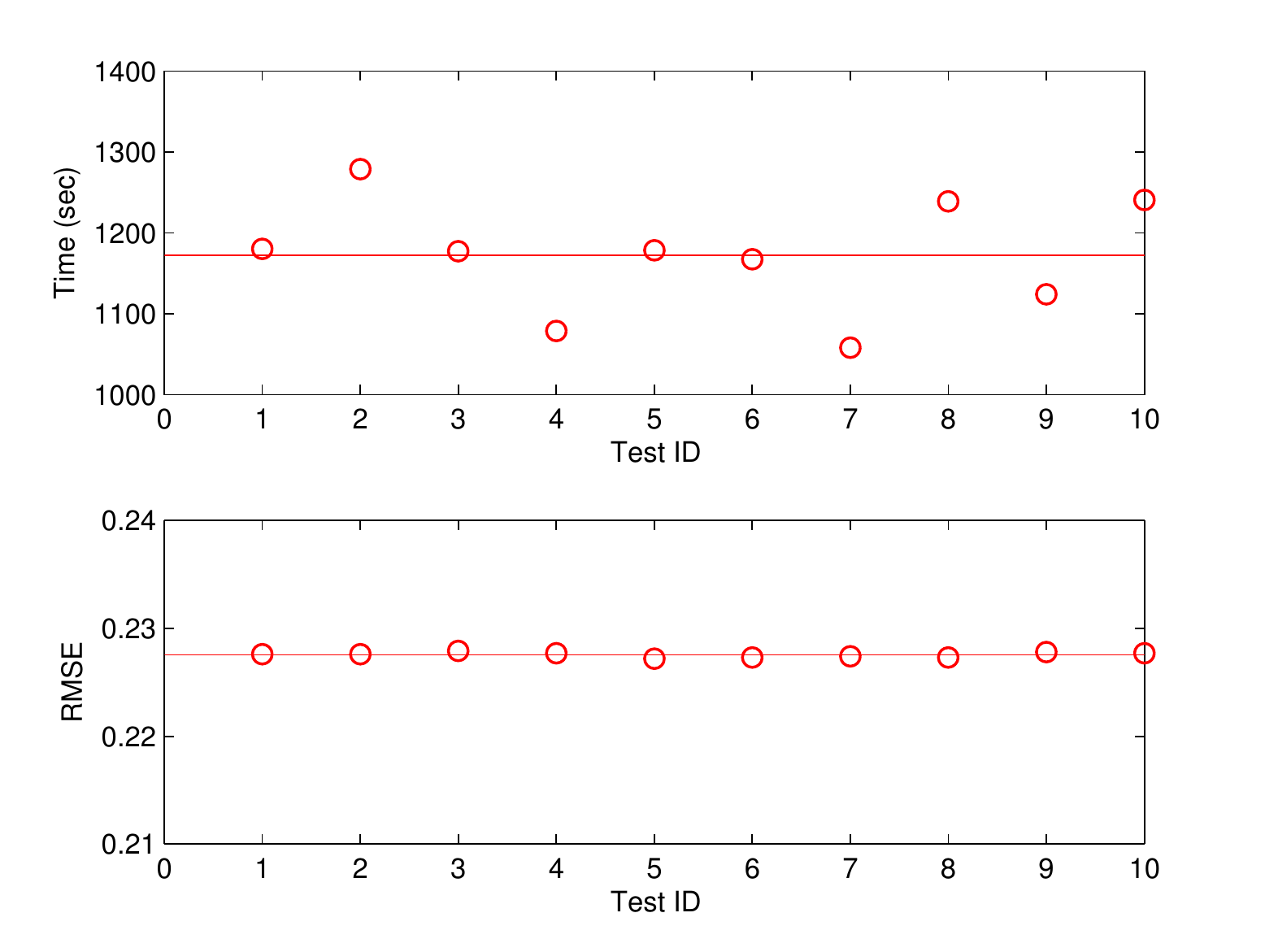}}
\end{minipage}   
\begin{minipage}[b]{0.45\textwidth}  
\subfigure[$1000 \times 1000 \times 1000$]{\label{subfig:scalable1000}\includegraphics[scale=0.40]{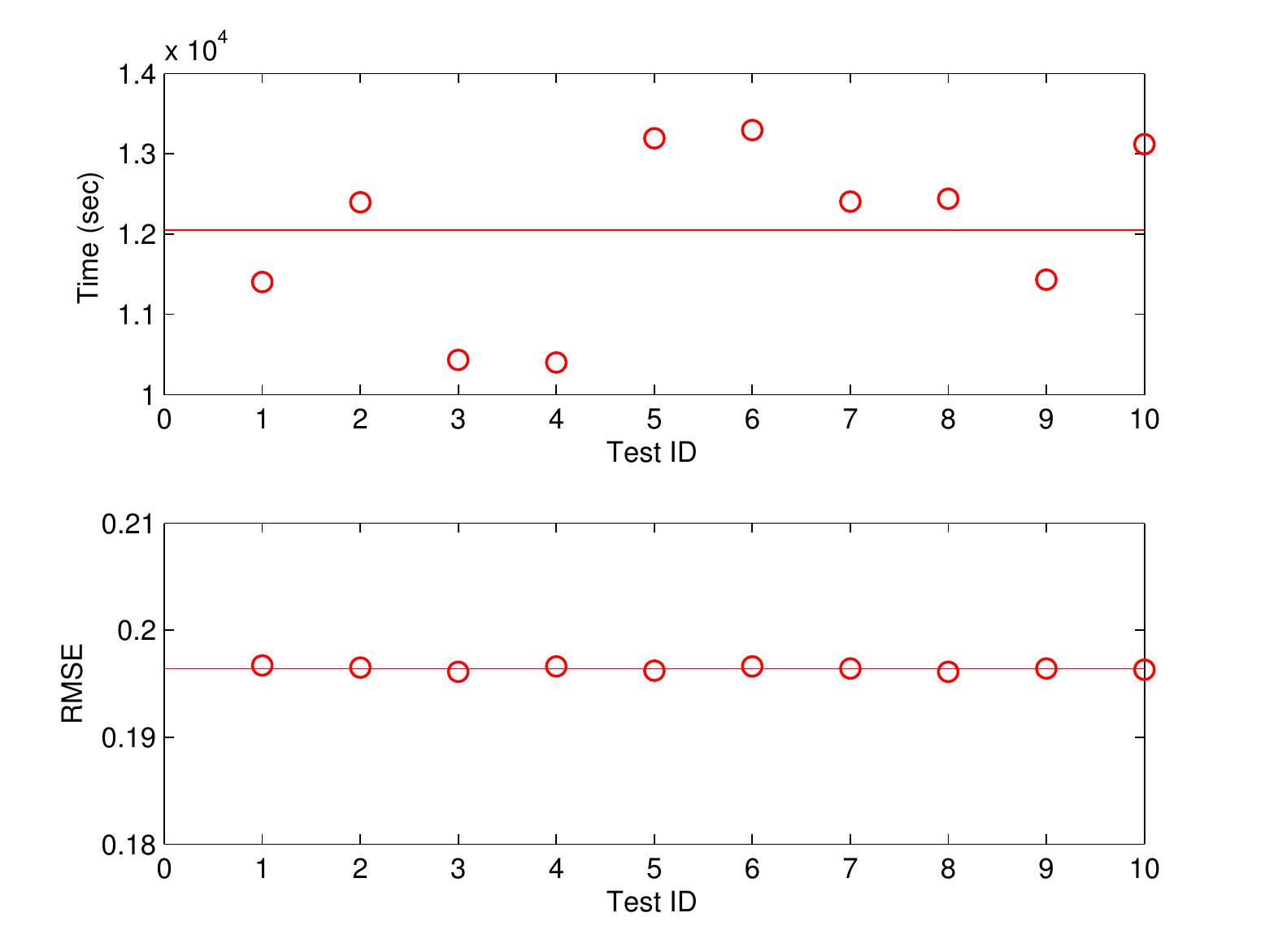}}
\end{minipage}
\caption{Results of our algorithm for large-scale problems. The means are shown as solid lines.}
\label{fig:sparseVB}
\vspace*{-5mm}
\end{figure}

To test the scalability of the PLTF-VB and GCTF-VB methods, we have conducted experiments on tensor completion problem to demonstrate that time complexity of the modeling framework is $O(N)$ for sparse datasets. We consider two situations in these experiments: (i) $500 \times 500 \times 500$ three-way array with $99\%$ missing data (1.25 million known values), and (ii) $1000 \times 1000 \times 1000$ three-way array with $98.75\%$ missing data (12.5 million known values). We have used CP tensor factorization model with $R=5$ components to generate data, then added $20\%$ random Gaussian noise. We have then fitted a CP model using KL-divergence as loss function and used the extracted CP factors to reconstruct the data. Figure~\ref{fig:sparseVB} shows the \emph{average tensor completion performance of 10 independent runs} in terms of RMSE score. In the $500 \times 500 \times 500$ case, all ten problems have been solved with an RMSE score around 0.23, with computation times ranging between 1000 and 1300 seconds and in the $1000 \times 1000 \times 1000$ case, all ten problems are also solved with an RMSE score around 0.20. The computation times have ranged from 10000 to 13000 seconds, approximately 10 times slower than the $500 \times 500 \times 500$ case, which has 10 times more non-missing entries.
% Scalability
% Second Set of Experiments
\paragraph{\textbf{Hyperparameter Selection:}} We observe that hyperparameter adaptation is crucial for obtaining good prediction performance. In our simulations, results for PLTF-VB without hyperparameter adaptation were occasionally poorer than the PLTF-EM estimates. We set both shape $A$ and scale $B$ hyperparameters same for all components $Z_{1:3}$. We tried several number of different values for hyperparameters to obtain the best prediction results under missing data case. Figure~\ref{fig:compHyperparameters} shows the comparison of three different hyperparameter settings; $A=0.5,B=10$, $A=10,B=10$ and $A=100,B=1$ in terms of link prediction performance. As we can see, we obtain best result when initialising the shape hyperparameter $A=0.5$ and scale hyperparameter $B=10$ for all settings of missing data. So, we use these values of hyperparameter $A$ and $B$ for the following experiments in section~\ref{sec:LPexperiments}. In addition, we obtain that when we set $A<1$ and $B>10$, we get better results.    

% Third Set of Experiments
% -------------------------------------------------------------------------------------------------------------------
\subsection{Link Prediction}
\label{sec:LPexperiments}

We now compare the standard tensor factorization methods, i.e., PLTF-EM and GCTF-EM, with the proposed variational methods, i.e., PLTF-VB and GCTF-VB, on a missing link prediction task. 
%\paragraph{\textbf{Evaluation Metric:}}
In our experiments, we use Area Under the Receiver Operating Characteristic Curve (AUC) to measure the link prediction performance. Link prediction datasets are characterized by extreme imbalance, i.e., the number of links known to be present is often significantly less than the number of edges known to be absent. This issue motivates the use of AUC as a performance measure since AUC is viewed as a robust measure in the presence of imbalance \cite{StagerLT06}. The following results show the average link prediction performance of 10 independent runs in terms of AUC. 
\paragraph{\textbf{Results:}} We compare the performance of standard and variational approaches of PLTF and GCTF on both CP and Tucker tensor factorization models at different amounts of randomly unobserved elements. In these experiments, the incomplete tensor is factorized using either a CP or a Tucker model and the extracted factor matrices are used to construct the full tensor and estimate scores for missing links. For all cases, variational approach outperforms the standard approach clearly. Table~\ref{table:ComparisonAllCP} shows the time and accuracy performances of PLTF-EM, PLTF-VB, GCTF-EM and GCTF-VB methods for the CP model given in Equation(\ref{eq:CP}), when $\lbrace 60, 80, 90\rbrace$ of the data is missing. Based on these results, we can conclude two results: (i) the variational methods due to implicit self-regularization effect \cite{NakajimaS10}, perform better than the standard methods; (ii) coupled models outperform the single models in particular, when the percentage of missing data is high. 

\renewcommand{\arraystretch}{1.2}
\begin{table*}[ht!]
\begin{center}
%\hspace*{-8mm}
\scalebox{0.80}{
%\begin{tabular}{| L{15mm} | C{15mm}  |  C{10mm} || c | c || c | c || c | c |}
\begin{tabular}{| c | c | c || c | c || c | c || c | c |}
\hline  &   \multicolumn{2}{c ||}{}   & \multicolumn{2}{c ||}{60\% missing} & \multicolumn{2}{c ||}{80\% missing} & \multicolumn{2}{c |}{90\% missing} \\ \cline{4-9} Dataset  & \multicolumn{2}{ c ||}{Algorithm}  &  AUC  &  Time(sec)  &  AUC  &  Time(sec)  &  AUC  &  Time(sec) \\ \hline
\multirow{4}{*}{UCLAF}    & \multirow{2}{*}{PLTF} &  EM  &  0.940 $\pm$ 0.004  &  1.69  &  0.867 $\pm$ 0.005  &  1.57  &  0.844 $\pm$ 0.005 &  1.43   \\ \cline{3-9}
&   &  VB  & \textbf{0.973 $\pm$ 0.002} & 2.12 &  \textbf{0.959 $\pm$ 0.003}  & 2.04  &  \textbf{0.917 $\pm$ 0.003}  &  1.93  \\ \cline{2-9}
&  \multirow{2}{*}{GCTF}  &  EM  & 0.917 $\pm$ 0.003 & 5.14  &  0.892 $\pm$ 0.004 & 5.08  & 0.869 $\pm$ 0.004 &  4.98   \\ \cline{3-9}
&   &  VB  &  \textbf{0.981 $\pm$ 0.001} &  6.19  &  \textbf{0.962 $\pm$ 0.001}  & 6.16 & \textbf{0.939 $\pm$ 0.002} & 6.01 \\ \hline \hline
% Comment Prediction
\multirow{4}{*}{\pbox{18mm}{Digg - \\ (Comment \\ Prediction)}} & \multirow{2}{*}{PLTF} &  EM  &  0.848 $\pm$ 0.003 &  582.54  &  0.829 $\pm$ 0.004 & 326.16  &  0.813 $\pm$ 0.004 &  167.98  \\ \cline{3-9}
&    &  VB   &  \textbf{0.917 $\pm$ 0.002}  &  787.94  & \textbf{0.897$\pm$ 0.003}  & 460.24 & \textbf{0.879 $\pm$ 0.003}  & 243.75 \\ \cline{2-9}
&  \multirow{2}{*}{GCTF}  &  EM  & 0.856 $\pm$ 0.002 &  1019.78  &  0.837 $\pm$ 0.003 &  598.22 &  0.824 $\pm$ 0.003 & 312.53  \\ \cline{3-9}
&    &  VB   &  \textbf{0.928 $\pm$ 0.001}  &  1251.40  &  \textbf{0.913 $\pm$ 0.002}  &  735.47  &  \textbf{0.891 $\pm$ 0.002}  &  452.99  \\ \hline \hline
% Digg Prediction
\multirow{4}{*}{\pbox{18mm}{Digg - \\ (Digg \\ Prediction)}} & \multirow{2}{*}{PLTF} &  EM  &  0.864 $\pm$ 0.005 &  277.11  &  0.845 $\pm$ 0.006 & 159.51  & 0.829 $\pm$ 0.006  & 89.54  \\ \cline{3-9}
&    &  VB   &  \textbf{0.935 $\pm$ 0.003}  &  340.20  &  \textbf{0.917 $\pm$ 0.003}  &  221.78  & \textbf{0.898 $\pm$ 0.004}  & 127.35  \\ \cline{2-9}
&  \multirow{2}{*}{GCTF}  &  EM  &  0.892 $\pm$ 0.003 &  473.42  &  0.870 $\pm$ 0.004 & 290.34  &  0.853 $\pm$ 0.004 &  168.51  \\ \cline{3-9}
&     &  VB   &  \textbf{0.961 $\pm$ 0.001}  &  537.54  &  \textbf{0.947 $\pm$ 0.001}  & 354.18  &  \textbf{0.923 $\pm$ 0.002}  & 211.15  \\ \hline
\end{tabular}}
\caption{AUC score (by \emph{'mean $\pm$ std'}) and time comparison of EM approaches and the proposed VB algorithms on various data sets with CP-tensor factorization model and different proportion of missing data. Results are averaged over 10 runs.}
\label{table:ComparisonAllCP}
\end{center}
\vspace*{-5mm}
\end{table*}

\begin{figure}
\hspace*{-15mm}
\begin{minipage}[b]{0.35\textwidth}  
\centering
\subfigure[GCTF-EM, 80\%]{\label{fig:compR2vsR20_EM}\includegraphics[scale=0.37]{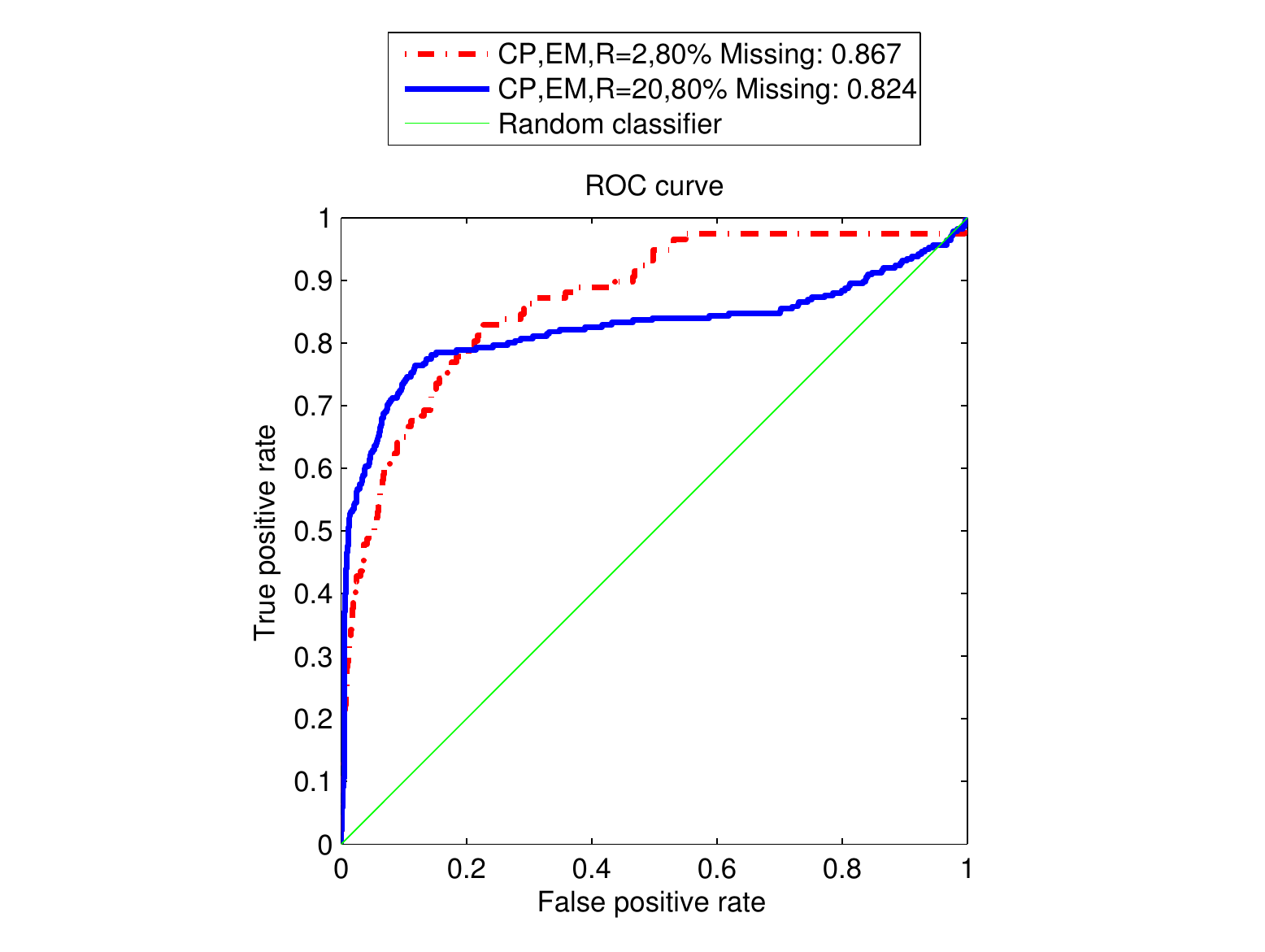}}
\end{minipage}
\begin{minipage}[b]{0.35\textwidth}  
\centering
\subfigure[GCTF-VB, 80\%]{\label{fig:compR2vsR20_VB}\includegraphics[scale=0.37]{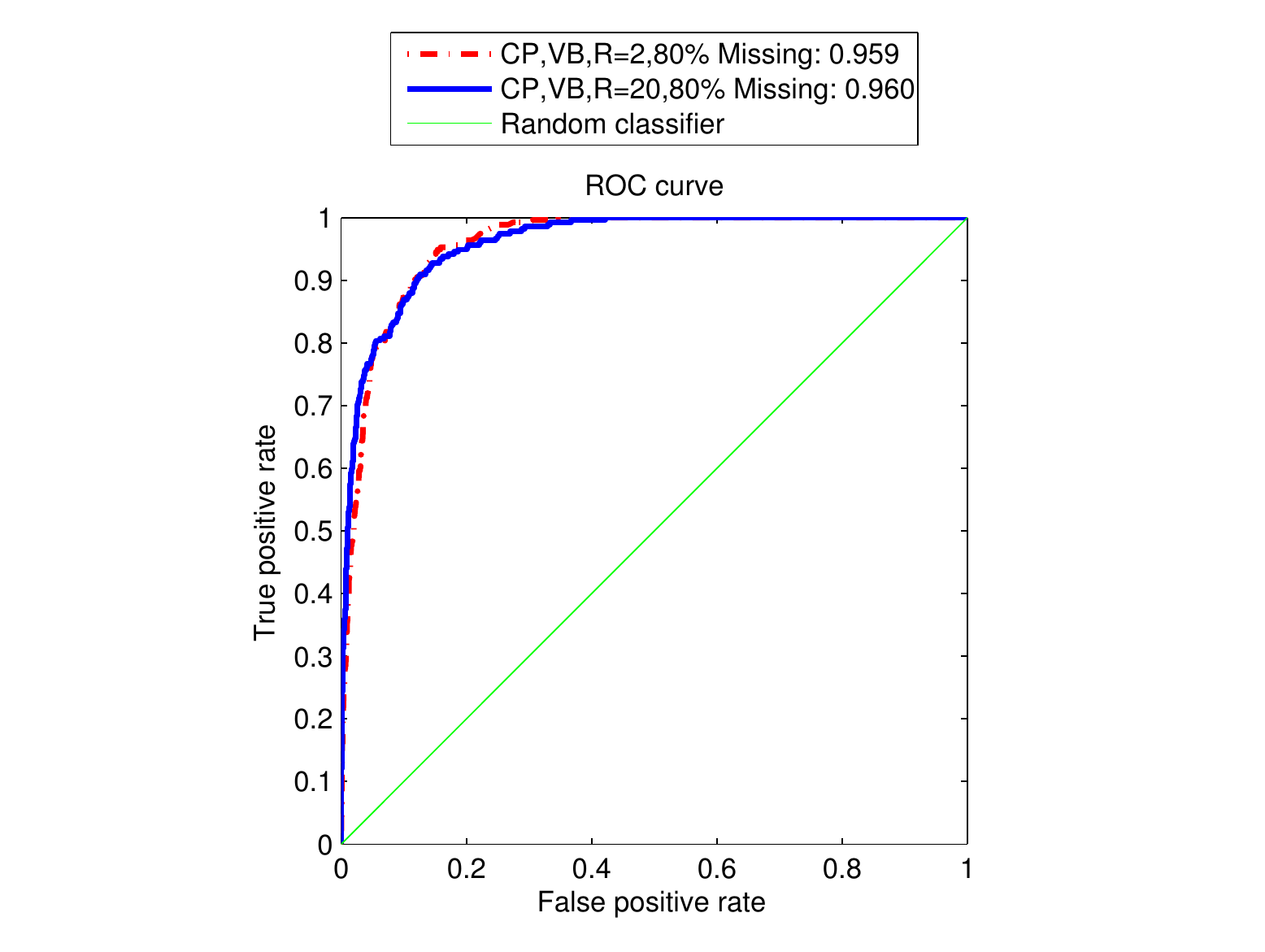}}
\end{minipage}
%
% Comparison of VB with different hyperparameters and different missing value rates
\begin{minipage}[b]{0.30\textwidth}  
\centering
\subfigure[GCTF-VB, 80\%]{\label{fig:compHyperparameters}\includegraphics[scale=0.39]{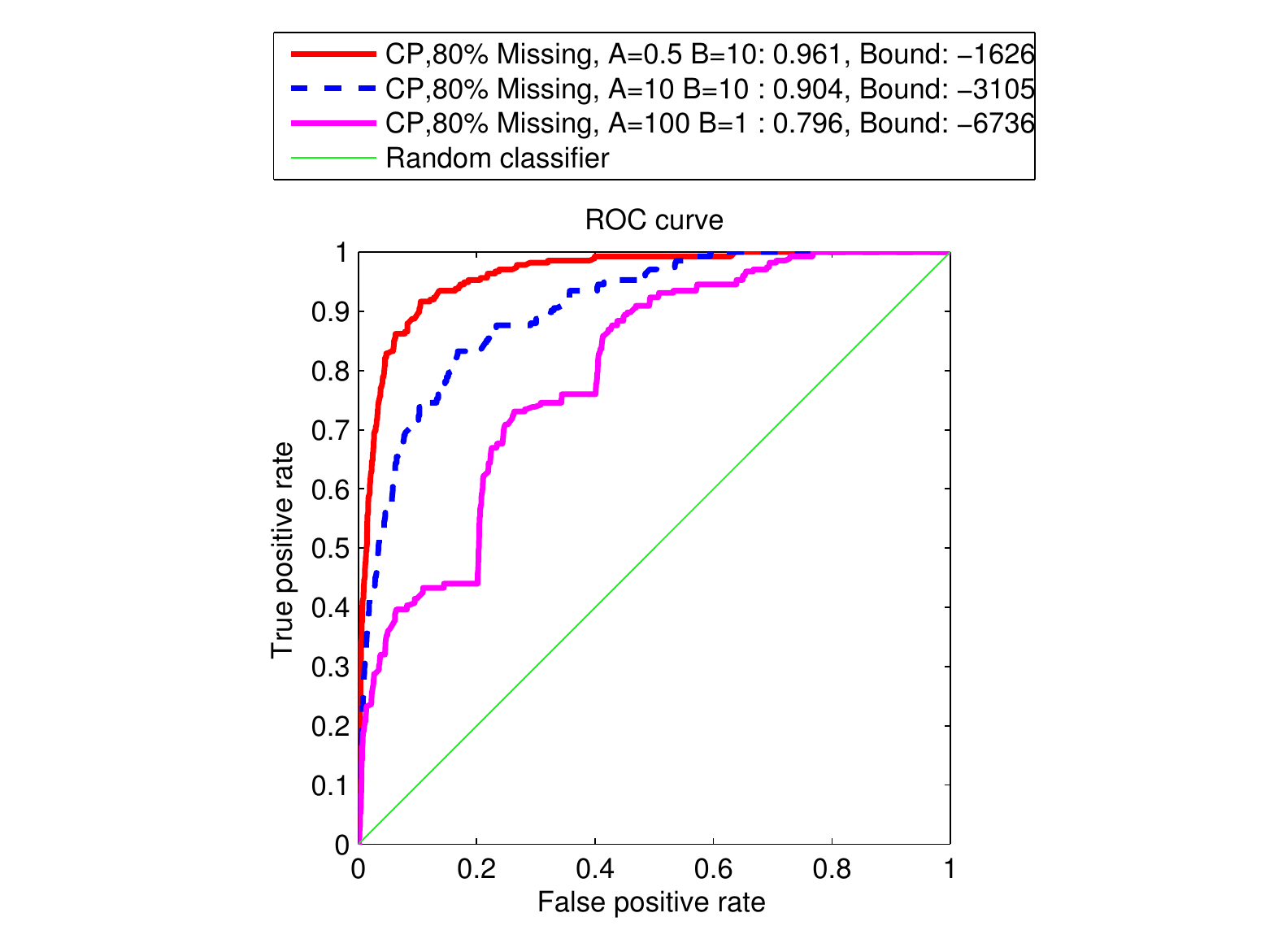}}
%\captionof{figure}{Effect of hyperparameter selection with CP model when R=2.}
\end{minipage}
\caption{Effect of model order on the performance of a) GCTF-EM and b) GCTF-VB approaches and c) Effect of hyperparameter selection for CP model when R=2.}
\vspace*{-5mm}
\end{figure}

Moreover, we study the performance of GCTF-EM and GCTF-VB in terms of robustness to model order selection. As model order increases, the prediction performance of GCTF-EM drops. This is as expected since GCTF-EM is prone to overfitting and the increase in model order causes an increase in the number of free parameters that, in turn, enlarges penalty term in GCTF-EM. On the other hand, the prediction performance of the variational approach is not very sensitive to the model order and is immune to overfitting since Bayesian approach alleviates over-fitting by integrating out all model parameters \cite{cemgil09-nmf}. We compare the prediction performances of GCTF-EM and GCTF-VB methods for the CP tensor model when the component number $R$ is equal to $2$ and $20$ and for different amounts of missing data, i.e., $\lbrace 40, 60, 80\rbrace$ of the data is missing. Figure~\ref{fig:compR2vsR20_EM} and Figure~\ref{fig:compR2vsR20_VB} demonstrate that when the model order increases, the prediction performance of PLTF-VB approach stays almost same; however, the prediction performance of PLTF-EM approach declines as expected.   
% Comparison of VB case with different R's with different missing value rates
% ----------------------------------------------------------------------

\section{Related Work}
\label{sec:related}

%In this section, we briefly introduce some of the related work in two categories: Bayesian inference for matrix and tensor factorizations and link prediction. 
Matrix and tensor factorization models are core components of collaborative filtering or link prediction and have attracted a large amount of research to date. We restrict our focus to probabilistic approaches.
In order to deal with the variational Bayesian matrix and tensor factorization problem, Ghahramani and Beal \cite{GhahramaniB00} provides a method that focus on deriving variational Bayesian learning in a very general form, relating it to EM, motivating parameter-hidden variable factorizations, and the use of conjugate priors. Shan et al. \cite{ProbTechreport} propose probabilistic tensor factorization algorithms, which are naturally applicable to incomplete tensors. First one is parametric probabilistic tensor factorization (PPTF), as well as a variational approximation based algorithm to learn the model and the second one is Bayesian probabilistic tensor factorization (BPTF) which maintains a distribution over all possible parameters by putting a prior on top, instead of picking one best set of model parameters. Cemgil \cite{cemgil09-nmf} describes a non-negative matrix factorization (NMF) in a statistical framework, with a hierarchical generative model consisting of an observation and a prior component. Starting from this view, he develops full Bayesian inference via variational Bayes or Monte Carlo.
Nakajima et al. \cite{NakajimaST10} propose a global optimal solution to variational Bayesian matrix factorization (VBMF) that can be computed analytically by solving a quartic equation and it is highly advantageous over a popular VBMF algorithm based on iterated conditional modes (ICM), since it can only find a local optimal solution after iterations. More recently, hierarchical Bayesian models for matrix co-factorization have been proposed \cite{Singh10,YooC11} to solve relational learning problems.

\section{Conclusions}
\label{sec:conc}

In this paper, we have investigated variational inference for PLTF and GCTF frameworks with KL cost from a full Bayesian perspective that also handles the missing data naturally. In addition, we develop a practical way without incurring much additional computational cost to EM approach for computing the approximation distribution and full conditionals. Our experiments demonstrate that the variational approach alleviates the overfitting better than the standard tensor factorization approaches and leads to the improved performance. 

% Bibliography
\clearpage
\bibliography{large_scale_nips2013}
\bibliographystyle{IEEEbib}

\end{document}